\documentclass[english,10pt,english,english,10pt,journal,english,compsoc]{IEEEtran}
\usepackage[LGR,T1]{fontenc}
\usepackage[latin9]{inputenc}
\usepackage{color}
\usepackage{array}
\usepackage{textcomp}
\usepackage{pmboxdraw}
\usepackage{multirow}
\usepackage{amsmath}
\usepackage{graphicx}
\usepackage{subscript}

\makeatletter

\ProvideTextCommand{\~}{LGR}[1]{\char126#1}

\newcommand{\lyxmathsym}[1]{\ifmmode\begingroup\def\b@ld{bold}
  \text{\ifx\math@version\b@ld\bfseries\fi#1}\endgroup\else#1\fi}

\providecommand{\tabularnewline}{\\}

\usepackage{url}

\@ifundefined{showcaptionsetup}{}{%
 \PassOptionsToPackage{caption=false}{subfig}}
\usepackage{subfig}
\makeatother

\usepackage{babel}
\begin{document}

\title{Detecting  Tiny Moving Vehicles in  Satellite Videos} 
\author{Wei Ao, Yanwei Fu, Feng Xu
\IEEEcompsocitemizethanks{  
\IEEEcompsocthanksitem  Wei Ao, Feng Xu are with Key Lab for Information Science of Electromagnetic Waves (MoE), Fudan University, Shanghai 200433, China. Email: \{wao16, fengxu\}@fudan.edu.cn
\IEEEcompsocthanksitem Yanwei Fu is with the School of Data Science Fudan University, Shanghai 200433, China. Email: yanweifu@fudan.edu.cn
}
\thanks{}
} 
\IEEEcompsoctitleabstractindextext{ 
\begin{abstract}  
In recent years, the satellite videos have been captured by a moving satellite platform. In contrast to consumer, movie, and common surveillance videos, satellite video can record the snapshot of the city-scale scene. In a broad field-of-view of satellite videos, each moving target would be very tiny and usually composed of several pixels in frames. Even worse, the noise signals also existed in the video frames, since the background of the video frame has the subpixel-level and uneven moving thanks to the motion of satellites.  We argue that this is a new type of computer vision task since previous technologies are unable to detect such tiny vehicles efficiently.  This paper proposes a novel framework that can identify the small moving vehicles in satellite videos. In particular, we offer a novel detecting algorithm based on the local noise modeling. We differentiate the potential vehicle targets from noise patterns by an exponential probability distribution. Subsequently, a multi-morphological-cue based discrimination strategy is designed to distinguish correct vehicle targets from a few existing noises further. Another significant contribution is to introduce a series of evaluation protocols to measure the performance of tiny moving vehicle detection systematically. We annotate a satellite video manually and use it to test our algorithms under different evaluation criterion. The proposed algorithm is also compared with the state-of-the-art baselines, and demonstrates the advantages of our framework over the benchmarks.

\end{abstract} 
} 
\maketitle

\section{Introduction\label{sec:Introduction}}

\begin{figure*}
\centering{}\subfloat[]{\includegraphics{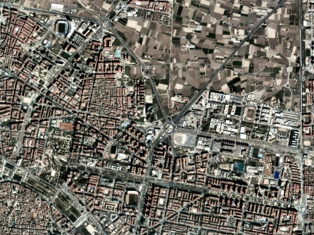}} \hspace{2cm}\subfloat[]{\includegraphics{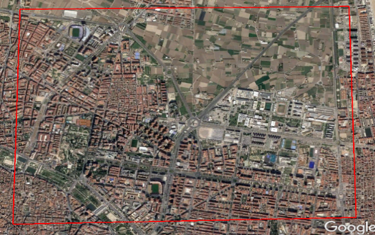}}\caption{An example of satellite video. (a) A frame of a satellite video of
Valencia, Spain. (b) Its corresponding optical map downloaded from
Google Earth.\label{fig:An-example-of} }
\end{figure*}

With the recent advanced in the earth observation (EO) technology,
satellite videos are captured by utilizing the optical sensors to
capture consecutive images from a moving satellite platform. The satellite
videos can enable many potential applications, such as city-scale
traffic surveillance, 3D reconstruction of urban buildings and quake-relief
efforts, \emph{etc}. For instance, Figure \ref{fig:An-example-of}
(a) shows that a frame of a satellite video of the Valencia city in
Spain. As visualized in Fig. \ref{fig:An-example-of} (b), corresponding
areal map of each video frame is about $3\times4$ square kilometers.
Satellite video can thus facilitate monitoring the dynamics scenes
of the city-scale. On the other hand, to efficiently supervise the
city-scale scene, one primary, and yet the critical task is to detect
and track the moving vehicles captured in satellite videos. However,
there is no previous technique for detecting the very tiny moving
vehicles in satellite videos, due to the following challenges.

\begin{figure}
\centering{}\subfloat[]{\includegraphics[scale=0.7]{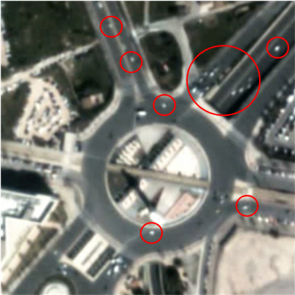}

} \hspace{0.5cm}\subfloat[]{\includegraphics[scale=0.7]{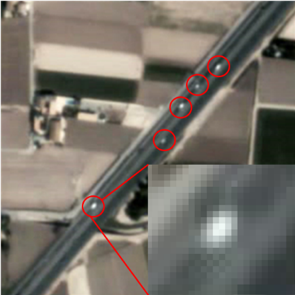}}\caption{The enlargements of two scenes in Fig. \ref{fig:An-example-of} (a),
where some vehicle targets are denoted with red circles.\label{fig:The-enlargement-of}}
\end{figure}

\textbf{In the video, vehicles are moving and very tiny. }In satellite
videos, only several pixels represent each vehicle. Thus essentially,
we have to detect tiny moving vehicles in satellite videos. Fig. \ref{fig:The-enlargement-of}
 shows two enlarged areas of the panorama in Fig. \ref{fig:An-example-of}~(a).
From these enlargements, a vehicle is only composed of \textit{several
pixels} without any distinctive color or texture. So, the state-of-the-art
detection algorithms, like deep learning, can easily overfit the training
vehicle data but fail to detect/describe the patterns of these moving
vehicles. We observe that the most robust features of these moving
vehicles come from their motion patterns which however are very easily
obscured and challenged by the background moving. 

\begin{figure}
\centering{}\subfloat[]{\includegraphics[scale=0.5]{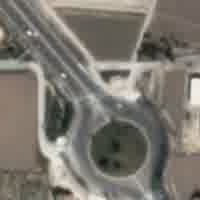}

} \hspace{0.5cm}\subfloat[]{\includegraphics[scale=0.5]{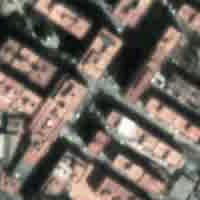}}
\hspace{0.5cm}\subfloat[]{\includegraphics[scale=0.5]{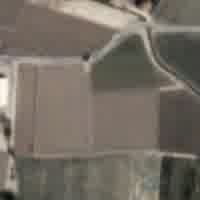}

} \hspace{0.5cm}\subfloat[]{\includegraphics[scale=0.5]{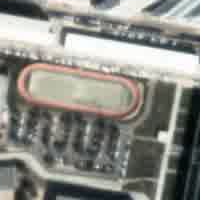}}
\caption{Varied parts of a frame of the satellite video. (a) roads, (b) buildings,
(c) vegetation and (d) a football field.\label{fig:Varied-parts-of}}
\end{figure}

\textbf{The frames of satellite videos cover a large-scale area and
provide a dynamic scenario.} In terms of the distances between camera
shot and observed objects, we have\emph{ }near-field, medium-field,
far-field surveillance videos, and the extremely far-field satellite
videos \cite{tain2011surveillance,chen2009farview}.  Not only broad
field-of-view does a satellite video provide but it also presents
a very complex background. As shown in Fig. \ref{fig:Varied-parts-of},
the visual content of satellite videos may include the roads, buildings,
vegetation and football field, \emph{etc}. Furthermore, it also has
varied traffic conditions as many as possible, \emph{e.g}. straight
arteries, intersections, and roundabouts, \emph{etc}. 

\begin{figure}
\begin{centering}
\subfloat[]{\includegraphics[scale=0.6]{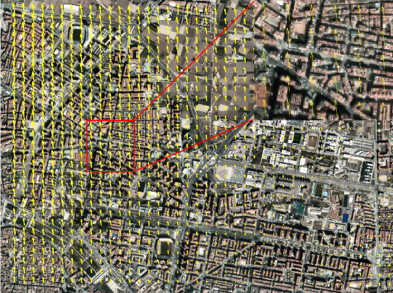}} \hspace{0.5cm}\subfloat[]{\includegraphics[scale=0.6]{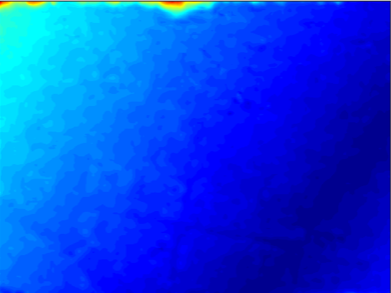}}
\par\end{centering}
\caption{Optical flow field which is obtained from frame 1 and 100 of the satellite
video. (a) shows the magnitude and orientation of the optical flow
field through vectors, while (b) further illustrates the magnitude
distribution of the optical flow filed. \label{fig:Optical-flow-field}}
\end{figure}

\textbf{The background of satellite videos presents sub-pixel-level
and uneven moving.} The optical flow field \cite{lucas1981iterative,efros2003action}
of the above satellite video is shown in Fig. \ref{fig:Optical-flow-field}.
It shows that the background is continuously moving, and the optical
flow field is very uneven. Even worse, the relative motion of the
satellite video is very complicated since intrinsically, the satellite
video frames are the 2D projection of a sophisticated 3D movement
of the satellite platform. On the other hand, since the satellites
are very far away from the earth plane, we can only observe very slow
moving among the consecutive video frames. Such slowly moving will
lead to small variants of the stationary pixels. Critically, we notice
that the moving of two successive satellite video frames is always
sub-pixel-level. And it also challenges the techniques of frame-by-frame
video stabilization and registration. Overall, one key difficulty
in detecting the tiny moving vehicles is to differentiate the background
motions from the moving vehicles; otherwise, the moving background
would negatively affect the detection of tiny moving vehicles.

This paper focuses on detecting and tracking the tiny vehicle of only
a few pixels in the satellite videos which is very hard to be identified
and easily affected by noise. The patterns of moving vehicles may
also be confused with the noise patterns which caused by the complex
moving backgrounds. Such noise patterns may result in regular moving
of stationary corners or edges, and thus further hinder the detection
of the tiny moving vehicles. 

To tackle the problems as mentioned above, we, for the first time,
propose a framework in addressing the challenging tasks of detecting
tiny moving vehicles in satellite videos. The whole framework is built
upon a series of statistical tools. In particular, we propose a motion
based detecting algorithm using a novel local modeling. We decompose
each frame into two parts, \emph{i.e.}, an original image and an additive
random 2D noise signal map. A probability distribution is used to
fit the noise patterns, which facilitates us to distinguish potential
vehicle targets. A local tactic is applied to address intra-variants
within a frame and discern inter-variants between frames, simultaneously.
Then, a region growing is designed, and a discrimination algorithm
based on multiple morphological cues is proposed, which can remove
other noise. The Kalman filter (KF) \cite{aru2002tracking} is further
used to track the vehicles. Extensive experiments are conducted on
the real-world satellite video dataset to evaluate our framework and
show the efficacy of the proposed models over the baselines.

The major contributions of this paper are fourfold:
\begin{enumerate}
\item To the best of our knowledge, the tasks of detecting moving tiny moving
objects are, for the first time, studied. To further study this problem,
we contribute the satellite video dataset, which has the labeled ground-truth
of tiny moving objects.
\item We propose a motion based detecting algorithm using a novel local
modeling. The noise pattern is modeled by probabilistic distributions.
\item A region growing algorithm is further designed and a discrimination
algorithm based on multiple morphological cues.
\item We, for the first time propose a set of evaluation protocols which
can systematically measure the algorithms of analyzing tiny moving
vehicles.
\end{enumerate}
The remainder of the paper is organized as follows. First, Sec. \ref{sec:Related-Work}
reviews some related work. Then, Sec. \ref{sec:Vehicles-Detection-and}
details the proposed algorithms, including the overall framework and
two major contributions, detecting and discrimination algorithms.
Subsequently, the used evaluation metrics and the proposed evaluation
algorithm is presented in Sec. \ref{sec:Metrics-and-Protocol}. Sec.
\ref{sec:Experiments-and-Discussion} shows experiments undertaken
on a satellite video. Finally, Sec. \ref{sec:Conclusion} concludes
the paper. 

\section{Related Work\label{sec:Related-Work}}

\subsection{Earth Observation Technology and Satellite Videos}

Nowadays many observation technologies have been developed and are
of enormous significance, including optical satellite images, space-borne
synthetic aperture radar (SAR) images and aerial videos. Such technology
plays a critical role in both civil and military area, such as the
city traffic system, maritime surveillance, aerial spy and battlefield
monitor, \emph{etc}. Both optical satellite image and space-borne
SAR can observe a large area in a high resolution. However, optical
satellite is very susceptible to different illumination and various
weather. Although SAR has the unique capability of earth imaging in
all-whether regardless of day and night \cite{ao2018detection}, the
SAR images are difficult to be interpreted \cite{zhou2016polarimetric,xu2016preliminary}.
Another weakness in understanding the optical satellite image and
SAR is that they cannot observe dynamics due to stationary imaging,
which narrows down their applications. 

The satellite videos have many advantages over the other conventional
videos, such as aerial videos captured by the unmanned aerial vehicle
(UAV). The aerial videos often suffer from undesirable dramatic motions
of platforms and have to resort to complex stabilization preprocessing
\cite{walha2015video}. Thus to track objects, image registration
has to be done at the first stage. One can then separate camera egomotion
from object motion \cite{jackson2010registering,guo2016joint,molina2014persistent}.
The aerial videos can only cover a small city scope, while our satellite
videos can easily supervise a city-scale scene. Furthermore, the new
legislation of the civil aviation safety had forbidden or restricted
the usage of UAV in many cities.

Recently, some commercial companies are able to have the satellites.
For example, Satellite Imaging Corporation (SIC) successfully launched
video satellites \textendash{} SkySat-1 and SkySat-2, on November
21, 2013 and July 8, 2014, respectively. Chang Guang Satellite Technology
CO., LTD (CGSTL) successfully launched two video satellites on October
7, 2015. Up to now, CGSTL has 8 on-orbit video satellites which are
a part of the ongoing Jilin No. 1 satellite constellation. The time
resolution of EO of Jilin No. 1 satellite constellation will be shortened
to half an hour when the constellation is constructed by 2020. 

Comparing with the conventional EO technologies. the satellite video
can cover the largest scope, and is very stable. First, we can understand
and forecast the dynamics of the earth. Second, a video satellite
can turn its lens towards the region of interest (ROI) all the time
through flying so long as this area is within the field-if-view of
the satellite, which has very good image quality. So, the satellite
videos are more stable compared with aerial videos. Third, a high
altitude of a video satellite results in a broader field-of-view,
which even covers a city-scale area. Another important issue should
be taken into account is that a video satellite is a free platform
which can record anywhere in the earth without any restrictions. 

\begin{figure*}
\centering{}\includegraphics[scale=0.3]{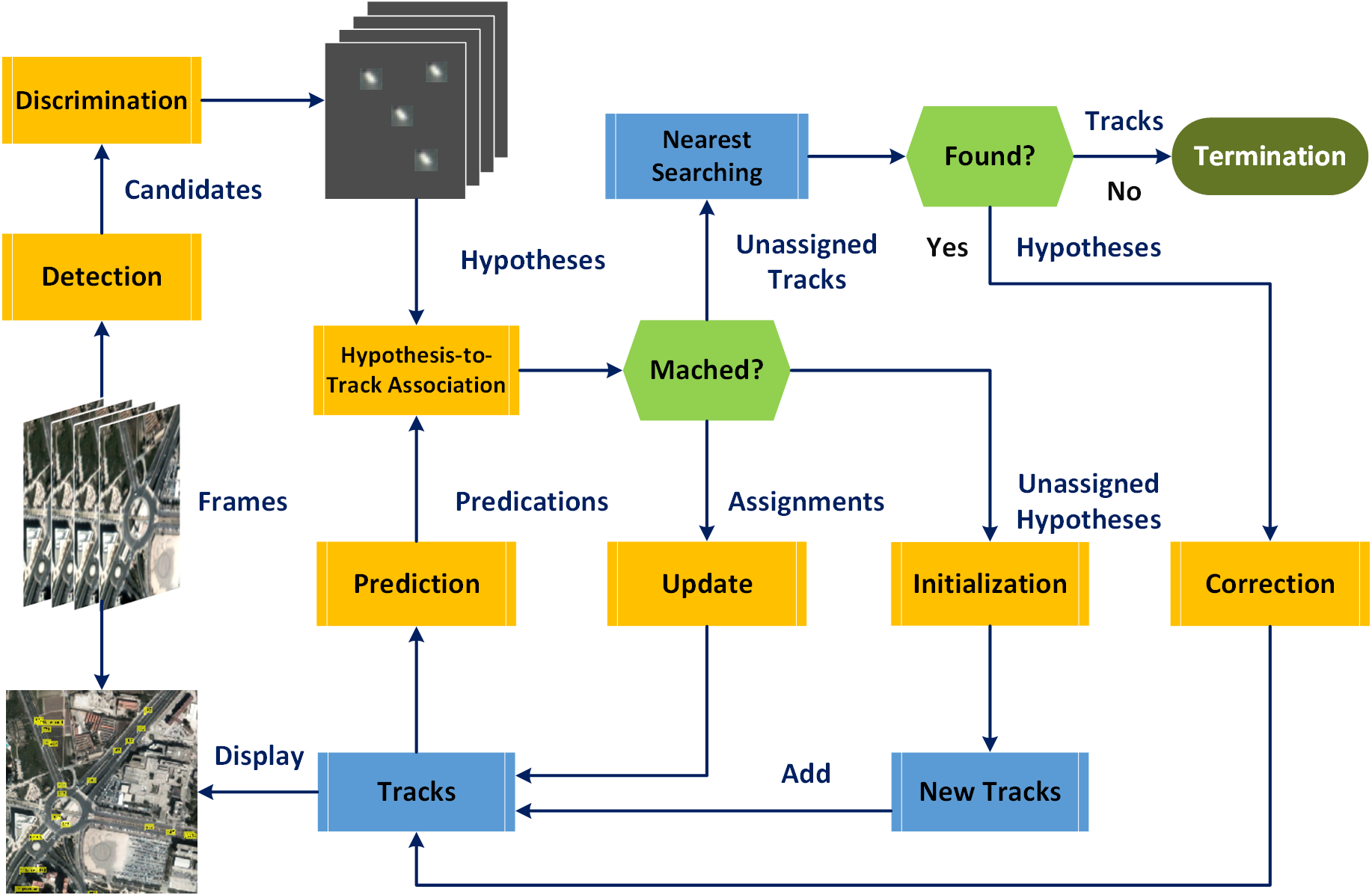}\caption{The overall framework of tiny vehicles detection.\label{fig:Fig.-9.-The} }
\end{figure*}

\subsection{Moving Target Detection and Tracking}

The moving target detection, can be taken as a special case of foreground
segmentation. Such tasks can be solved by the Gaussian Mixture Model
(GMM) \cite{jiang2014data,kaewtrakulpong2001improved,stauffer1999adaptive}
and the state-of-the-art ViBe \cite{barnich2009vibe,barnich2011vibe}
algorithm. GMM is a representative of parametric models, while the
ViBe tries a non-parametric method to describe the dynamic patterns
of a pixel. GMM utilizes a weighted mixture of Gaussian distributions
to model the pixel value varying over time. However, the dynamics
of pixel values may be not subjected to Gaussian distributions. In
most cases, we cannot use a definite parametric model to represent
the variance of pixel values. ViBe proposed a novel idea that some
pixel values in different time steps are regarded as the samples of
one space, in order to represent the patterns of the pixels. 

Those previous algorithms such as GMM and ViBe have several serious
drawbacks if we apply them to our tasks. First, they requires heavy
computational cost and resources in processing the satellite videos,
since the pixel-based modeling has heavy computational loadings. Second,
they are relative inefficiently in telling the differences between
the moving target and the ego-motion of the satellites. To this end,
we propose the noise model in isolating the moving background and
detecting the potential moving targets, simultaneously. Moreover,
the proposed tiny moving vehicle detecting is implemented in the spatiotemporal
domain. In terms of sub-pixel-level moving and the neighborhood similarity,
we model pixels of inter-frame differences spatially rather than temporally.

\begin{figure*}
\centering{}\subfloat[]{\includegraphics{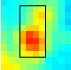}}\subfloat[]{\includegraphics{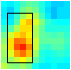}}\subfloat[]{\includegraphics{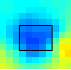}}\subfloat[]{\includegraphics{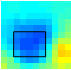}}\subfloat[]{\includegraphics{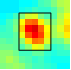}}\subfloat[]{\includegraphics{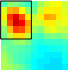}}\caption{Some cases of nearest searching, wherein the black rectangles denote
the tracking vehicles. (a), (c) and (e) are previous positions of
the vehicles, while (b), (d) and (f) are corresponding current positions.
Please note that it is hard to recognize vehicles by naked eyes in
original RGB images due to the low-contrast; so, the original images
are converted to colorful pseudo color images just for view.\label{fig:nearest-searching}}
\end{figure*}

\section{Methodology of Detecting Tiny Moving Vehicles \label{sec:Vehicles-Detection-and}}

The whole section is divided into the three subsections. Firstly,
we discuss the overall framework of proposed tiny moving vehicles
detection algorithms in Sec. \ref{subsec:Overview}. The proposed
framework is under Kalman filter (KF) tracking framework as shown
in Fig. \ref{fig:Fig.-9.-The}. These tiny moving vehicles are detected
in each local region in Sec. \ref{subsec:Motion-Based-Detection-Using}.
We subsequently propose the discrimination algorithm removing the
falsely detected components in Sec. \ref{subsec:Region-Growing-and}. 

\subsection{Overview of the Proposed Algorithms\label{subsec:Overview}}

\noindent \textbf{Key concepts}. Before fully developing our framework,
some key concepts are explained here.\textit{ Detection }or a\textit{
detector} is a potential-vehicle detecting procedure that embodies
the proposed local tactic and noise modeling algorithm. \textit{Candidates}
are outputs of the detection that are composed of true vehicles and
some noises. \textit{Discrimination} or a \textit{discriminator} is
a distinguishing procedure between true vehicles and existing noises,
including the proposed region growing and multi-morphological-cue
based discrimination algorithms.\textit{ Hypotheses} are outputs of
the discrimination that are composed of true vehicles and a few noises.
In addition, the final outputs of the detecting and tracking framework
are also defined as \textit{hypotheses}. The \textit{State} is a vector
that includes position, velocity and acceleration of a vehicle in
a time step. The \textit{Track} is a sequence of states of a vehicle
in temporal domain. Each track is marked with an unique ID and assigned
with a KF. \textit{Association} is a matching procedure that meets
the minimum cost. \textit{Prediction} is a current position of a track
that is inferred by its KF. 

The widely-used object tracking methods include Kalman filter, particle
filter \cite{aru2002tracking}, and mean shift \cite{app_density}.
As shown in Fig. \ref{fig:Fig.-9.-The}, our whole vehicle tracking
pipeline is based on Kalman filter, which is one of the most classical
tracking algorithms. In Fig. \ref{fig:Fig.-9.-The}, the KF is the
central module of the processing framework, which includes several
interactive branches, as follows.

\vspace{0.05in}

\noindent \textbf{(1) Initialization.} Initialization is to determine
the initial state of a track. The hypotheses of current frame and
previous frame are associated using Hungarian algorithm \cite{Miller1997Hungary,stiefelhagen2006clear}.
So, we can derive their velocities and positions, and their initial
accelerations are regarded as zero.

\vspace{0.05in}

\noindent \textbf{(2) Prediction.} The current state of one vehicle
tracked can be inferred in term of the previous observation. 

\vspace{0.05in}

\noindent \textbf{(3) Hypothesis-to-Track Association. }The discriminator
yields hypotheses, while the tracks yield predictions. In this stage,
hypotheses are matched with predictions in order to meet minimum cost.
Here, the cost between a hypothesis and a prediction is defined as
Euclidean distance. Hungary algorithm is employed to derive an optimal
association between hypotheses and predictions. Hypothesis-to-track
association yields assignments, unassigned tracks and unassigned hypotheses.
\textit{Assignments} are optimal matches. \textit{Unassigned tracks}
are those tracks that do not successfully with any hypotheses, likewise
\textit{unassigned hypotheses} do. Then, assignments are utilized
to update the stages; unassigned tracks are further processed in the
nearest searching stage; unassigned hypotheses are used to initialize
new tracks.

\vspace{0.05in}

\noindent \textbf{(4) Update.} The state vectors of hypotheses of
the assignments are used to update the state of their corresponding
KFs. 

\vspace{0.05in}

\noindent \textbf{(5) Nearest Searching, correction and termination.}
We do not simply discard the unassigned tracks because their corresponding
hypotheses may be missed by the detector or the discriminator. So,
the nearest searching strategy is applied to find out whether there
exists a connected region which resembles the tracking vehicle around
the previous position of the vehicle. The matching is resort to structural
similarity index (SSIM) \cite{wang2004image}. If the similar region
is found, the track is updated; otherwise the track is terminated.
The nearest searching using SSIM is illustrated in Fig. \ref{fig:nearest-searching}.
(a), (c) and (e) are from the previous frame where the vehicles are
marked by black rectangles, while (b), (d), and (f) are from the current
previous where the black rectangles denote the results of nearest
searching. These experimental results demonstrate the efficiency of
such a nearest searching strategy. 

\begin{figure*}
\centering{}\includegraphics[scale=0.6]{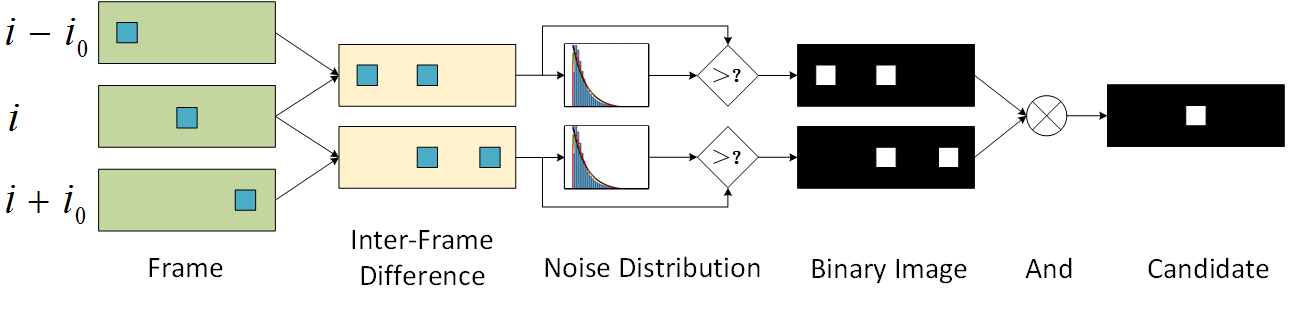}\caption{Flow diagram of the motion-based detection algorithm via noise modeling.\label{fig:Flow-diagram-of}}
\end{figure*}

\noindent  We will fully explain the above four steps of detecting
the tiny moving objects in the next few sections. Once the objects
are detected, the Kalman filter is adopted to fit the motion of these
vehicles, which is the optimal solution in the linear and Gaussian
situations. Although the motions of vehicles in real world are very
complex, from an approximate viewpoint, a non-linear procedure can
be decomposed into a series of linear procedures. So, KF is a simple
but effective tool to measure, predict and track the motion of a moving
vehicle. The evolution function of the system is defined as

\begin{equation}
\mathbf{x}_{i}=\mathbf{F}_{i}\text{\ensuremath{\cdot}}\mathbf{x}_{i-1}+\boldsymbol{v_{i}}\label{eq:9}
\end{equation}
where $\boldsymbol{x}\boldsymbol{_{i}}$, $\boldsymbol{F_{i}}$ and
$\boldsymbol{v_{i}}$ denote state vector, evolution matrix and procedure
noise vector, respectively, and the subscript $i$ indicates the time
step of a frame. Position, velocity and acceleration of a vehicle
constitute $\boldsymbol{x}\boldsymbol{_{i}}$, namely

\begin{equation}
\mathbf{x}_{i}=[x,y,v_{x},v_{y},a_{x},a_{y}]^{T}\label{eq:10}
\end{equation}
Without loss of generality, we assume that vehicle targets move in
a constant acceleration and straight line during each fixed interval.
So, the evolution matrix $\boldsymbol{F_{i}}$ can be written as 
\begin{equation}
\mathbf{F}_{i}=\left[\begin{array}{cccccc}
1 & 0 & \tau & 0 & \tau^{2}/2 & 0\\
0 & 1 & 0 & \tau & 0 & \tau^{2}/2\\
0 & 0 & 1 & 0 & \tau & 0\\
0 & 0 & 0 & 1 & 0 & \tau\\
0 & 0 & 0 & 0 & 1 & 0\\
0 & 0 & 0 & 0 & 0 & 1
\end{array}\right]\label{eq:11}
\end{equation}
The measurement function can be written as

\begin{equation}
\boldsymbol{y_{i}}=\mathbf{H}_{i}\cdot\mathbf{x}_{i}+\mathbf{n}_{i}\label{eq:12}
\end{equation}
where $\mathbf{y}_{i}$, $\mathbf{H}_{i}$ and $\mathbf{n}_{i}$ denote
measurement vector, measurement matrix and measurement noise, respectively.
The definition of $\boldsymbol{H_{i}}$ in this study is 

\begin{equation}
\mathbf{H}_{i}=\left[\begin{array}{cccccc}
1 & 0 & 0 & 0 & 0 & 0\\
0 & 1 & 0 & 0 & 0 & 0
\end{array}\right]\label{eq:13}
\end{equation}
Assuming that a set of measurements is obtained, \emph{i.e.} $\mathbf{y}_{1:i}=\left\{ y_{k}\lyxmathsym{\textSFxi}k=1,2,\cdots,i\right\} $,
KF recursively derives posterior PDF of the state vector \textbf{$\boldsymbol{x_{i}}$}
via Bayesian theorem, \emph{i.e}. 

\begin{equation}
p\left(\mathbf{x}_{i}\mid\mathbf{y}_{1:i}\right)=\frac{p\left(\mathbf{y}_{i}\mid\mathbf{x}_{i}\right)p\left(\mathbf{x}_{i}\mid\mathbf{y}_{1:i-1}\right)}{p\left(\mathbf{y}_{i}\mid\mathbf{y}_{1:i-1}\right)}\label{eq:15}
\end{equation}

\subsection{Motion-Based Detection Using Local Noise Modeling\label{subsec:Motion-Based-Detection-Using}\textcolor{red}{}}

Inter-frame difference is a conventional but effective tool to discerns
changes between two frames. In contrast tos pixel-based ViBe and GMM,
it has two notable merits: high efficiency and low memory consuming.
However, traditional inter-frame difference \cite{xiao2010vehicle}
is based on a predefined threshold to separate moving pixels and background.
Specifically, the grey-level value difference image is converted to
a binary image wherein ones denote moving pixels, while zeros denote
stationary background pixels. This procedure is termed as \textit{Binarization}
in our paper. Essentially, binarization differentiates moving pixels
from the whole inter-frame difference image. But a fixed binarization
threshold cannot be adapted to large-scale intro-variant scenarios
of satellite videos.

In order to address the aforementioned challenges, we propose a \textit{local
tactic} and a novel \textit{detecting method}. It is conceptualized
as motion-based detection using local noise modeling. An adaptive
binarization is derived by noise modeling in each local area, dealing
with the variances of local area by the neighborhood similarity. Motion-based
means inter-frame difference processing to search moving pixels.
\begin{itemize}
\item \textbf{Local~tactic.} A local tactic is designed to tackle the dramatic
intro-variance within a frame. Specifically, a 2D rasterizing is implemented
along the vertical and horizontal directions in a frame. The original
frame is converted into paved local areas. The size of a local is
empirically set as $30\times30$ square pixels. For one thing, a local
area has much lower degree of heterogeneity than the whole frame.
For another, it integrated local information to reduce the interference
of the moving background. It greatly facilitates the following detecting. 
\item \textbf{Detecting~method.}\textit{ }The proposed detecting method
is composed of four stages as shown in Fig. \ref{fig:Flow-diagram-of},
(1) \emph{deriving inter-frame difference images}, (2) \emph{estimating
noise distribution}, (3)\emph{ binarization} and (4)\emph{ logical
AND} operations to finally get the detection results. Our major contribution
is step (2) to estimate noise distribution which can yield a adaptive
binarization threshold of each local area in step (3). Each step details
as follows.
\end{itemize}
\begin{figure}
\centering{}\includegraphics{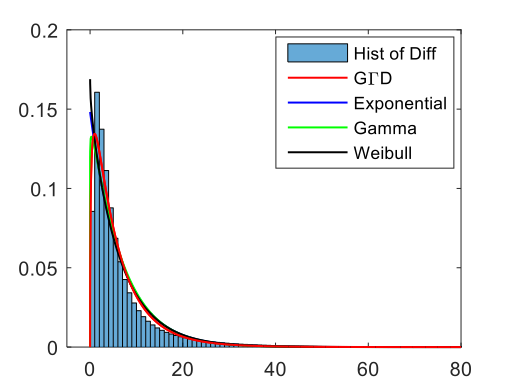}\caption{Amplitude histogram of noises and some fitted probabilistic distributions.
Note that \textquotedbl{}Histogram\textquotedbl{} and \textquotedbl{}Difference\textquotedbl{}
are abbreviated as \textquotedbl{}Hist\textquotedbl{} and \textquotedbl{}Diff\textquotedbl{},
respectively.\label{fig:Amplitude-histogram-of}}
\end{figure}

\begin{table}
\centering{}%
\begin{tabular}{c|c|c|c|c|c}
\hline 
Distance  & Exponential  & Gamma & Weibull  & G$\Gamma$D & Frame\tabularnewline
\hline 
\hline 
KL & 0.0959  & 0.0914  & 0.0919  & 0.0544 & \multirow{2}{*}{50}\tabularnewline
\cline{1-5} 
KS & 0.0959  & 0.1018  & 0.0891 &  0.0813 & \tabularnewline
\hline 
KL & 0.0864  & 0.0812 & 0.0862 &  0.0579 & \multirow{2}{*}{100}\tabularnewline
\cline{1-5} 
KS & 0.0875  & 0.0988  & 0.0896  & 0.0865 & \tabularnewline
\hline 
KL & 0.0846 & 0.0800 & 0.0845  & 0.0531 & \multirow{2}{*}{500}\tabularnewline
\cline{1-5} 
KS & 0.0854  & 0.0964  & 0.0859  & 0.0816 & \tabularnewline
\hline 
\end{tabular}\caption{KL and KS Distance of Some Noise Probability Models.\label{tab:KL-and-KS}
The smaller values, the better performance.}
\end{table}

\vspace{0.05in}

\noindent \textbf{(1) Deriving Inter-frame difference images.} Inter-frame
difference is a trivial operation. Here, we provide a novel viewpoint
on inter-frame difference images. By taking the frame as a 2D signal
consisting of original optical signal and additive random noise, \emph{i.e}.
\begin{equation}
G_{i}\left(x,y\right)=g_{i}\left(x,y\right)+n_{i}\left(x,y\right)\label{eq:1}
\end{equation}
 \noindent where $G_{i}\left(x,y\right)$ denotes the grey-level
value of pixel $\left(x,y\right)$ in frame $i$, since the gray-level
images\footnote{The RGB frames will be converted to grey-level images. }
are more common in satellite videos. $g_{i}(x,y)$ denotes the original
amplitude of the pixel $\left(x,y\right)$ in frame $i$, while $n_{i}(x,y)$
denotes the corresponding noise signal. Accordingly, the absolute
inter-frame difference of two registered frames can be regarded as
a set of random noise, \emph{i.e.,}

\begin{align}
D_{i,i+k}(x,y) & =\left|G_{i}(x,y)-G_{i+k}(x,y)\right|\label{eq:26}\\
= & \left|n_{i}(x,y)-n_{i,i+k}(x,y)\right|\label{eq:27}
\end{align}

\noindent where $D\left(\cdot\right)$ denotes absolute inter-frame
difference, $k$ denotes the \textit{k} frames interval. From Eq.
(\ref{eq:27}), the inter-fame difference image signal is only corresponding
with noises when two frames are registered. However, there still exists
some outliers. These outliers are composed of tiny moving vehicles
and non-vehicle targets. Thus the next issue is how to differentiate
these outliers from random noises. 

\vspace{0.05in}

\noindent \textbf{(2) Estimating noise distribution.} Detecting the
pattern of tiny moving vehicles is a challenge, since noise patterns
will blur the underlying patterns of the tiny moving vehicles. Thus
in this step, the key idea is to fit the noise patterns, namely $D_{i,i+k}(x,y)$
in Eq. (\ref{eq:26}), by the probabilistic distributions. 

Intuitively, the value differences of the same pixel at two consecutive
frames should approximate zero, while the value differences of the
pixels of noise patterns, or moving vehicles should be larger than
zero.  Figure \ref{fig:Amplitude-histogram-of} shows the histogram
of the value differences of pixels of two consecutive frames. The
amplitude histogram of noises exhibits notable regulations, like smooth
decaying and a heavy tail. In the pattern of noise, true noise pixels
are \textit{inliers}, and the other pixels are \textit{outliers}.
Thus the heavy tail of Fig. \ref{fig:Amplitude-histogram-of} should
be corresponding to the outliers.  

The probabilistic distribution is adopted to fit the histogram and
derive a binary threshold given a probability. Thus several widely-used
heavy-tail distributions, such as exponential distribution, Gamma
distribution, Weibull distribution and generalized Gamma distribution
(G$3\Gamma$D) are tested and compared in Fig. \ref{fig:Amplitude-histogram-of}.
Quantatively, Kullback-Leibler (KL) distance \cite{bishop2006prml}
and Kolmogorov-Smironv (KS) distance, (also known as Kolmogorove-Smirnov
test \cite{press1996numerical}), are introduced to quantify the fitting
performance of different distributions, as shown in Table \ref{tab:KL-and-KS}.
The smaller scores of KL and KS distance indicate the better results
for fitness. 

The quantitative experiment further proves the above hypothesis. It
also shows that the three-parameter distribution, G$\Gamma$D
with higher degree of freedom (DoF), outperforms the other distributions.
Alternatively, the one-parameter distribution \textendash{} exponential
distribution, also fit the noise distribution very well. Nevertheless,
the parameter estimation of G$\Gamma$D is more difficult, higher
computational load and more time consuming than exponential distribution.
To make a balance between accuracy and computational load, exponential
distribution is adapted to fit noises; and the cumulative density
function (CDF) is 

\begin{equation}
c_{E}(x;\lambda)=\begin{cases}
1-\exp(-\lambda x) & x>0\\
0 & x\leq0
\end{cases}\label{eq:3}
\end{equation}

\vspace{0.05in}

\noindent \textbf{(3) Binarization.} Once we fit the distribution
of pixel value differences of consecutive frames, we can utilize an
adaptive threshold of binarization to determine the outliers. In particular,
we introduce a predefined probability ($p_{fa}$) to derive the binarization
threshold, namely \cite{ao2018detection}
\begin{equation}
th=c_{E}^{-1}(1-p_{fa};\lambda)\label{eq:4}
\end{equation}
 \noindent where $c_{E}^{-1}(\cdot)$ denotes the inverse function
of the distribution in Eq (\ref{eq:3}). We set $p_{fa}$ as $5\times10^{-2}$
here. If the pixel value difference is bigger than the binarization
threshold $th$, this pixel would be taken as an outlier.\textcolor{blue}{{}
}By virtue of such a binarization algorithm, we turn the original
images into a binary image: inliers have zero pixel values, outliers
are ones. 

\vspace{0.05in}
\begin{figure}
\begin{centering}
\subfloat[]{\includegraphics{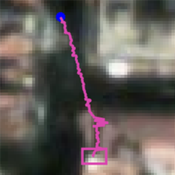}} \hspace{0.5cm}\subfloat[]{\includegraphics{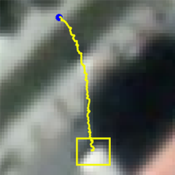}}\hspace{0.5cm}\subfloat[]{\includegraphics{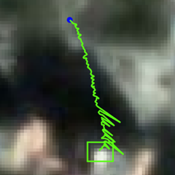}}\hspace{0.5cm}\subfloat[]{\includegraphics{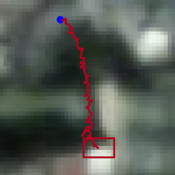}}
\par\end{centering}
\caption{Visualization of regular noises: the stationary corners or edges.
The rectangles denotes starting locations, while the filled blue points
are their terminal positions.\label{fig:Pseudo-motions}}
\end{figure}

\noindent \textbf{(4) Logical AND.} These outliers comprise the vehicles
and other noise. In the binary difference image, a true vehicle target
exhibits as two symmetrical blobs. One blob indicates the current
position, and another indicates the previous or future position. We
derive the intersection of two binary inter-frame difference images\footnote{Explicitly, the difference between frame \textit{$i$ }and frame $i-i_{0}$
vs. the difference between frame\textit{ $i$ }and the frame $i+i_{0}$,
wherein $i_{0}$ is set as 10.} to determine the current positions of vehicles. It is a Boolean operation
that only one and one yield one, which is named as, ``\textit{Logical
AND}''. In addition to eliminating ambiguities, logical AND also
reduces the existing noises due to their random appearing.

\subsection{Region Growing and Multi-Morphological-Cue Based Discrimination\label{subsec:Region-Growing-and}\textcolor{red}{{}
}}

\begin{figure*}
\begin{centering}
\subfloat[]{\includegraphics{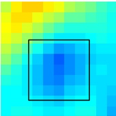}}\hspace{0.5cm}\subfloat[]{\includegraphics{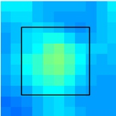}}\hspace{0.5cm}\subfloat[]{\includegraphics{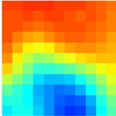}}\hspace{0.5cm}\subfloat[]{\includegraphics{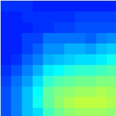}}
\par\end{centering}
\begin{centering}
\subfloat[]{\includegraphics{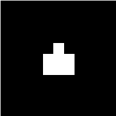}}\hspace{0.5cm}\subfloat[]{\includegraphics{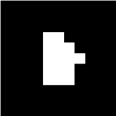}}\hspace{0.5cm}\subfloat[]{\includegraphics{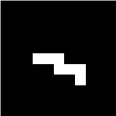}}\hspace{0.5cm}\subfloat[]{\includegraphics{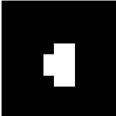}}
\par\end{centering}
\centering{}\subfloat[]{\includegraphics{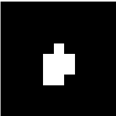}}\hspace{0.5cm}\subfloat[]{\includegraphics{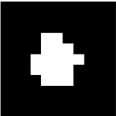}}\hspace{0.5cm}\subfloat[]{\includegraphics{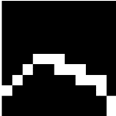}}\hspace{0.5cm}\subfloat[]{\includegraphics{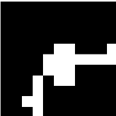}}\caption{Detecting and region growing results of vehicle targets and a noise.
(a) - (d) are pseudo color images converted from original RGB images
for view, where real vehicles are marked with black rectangles, and
(c), (d) are falsely detected edges or corners of buildings. (e) -
(h) are corresponding foregrounds generated by the detector. (i) -
(l) are reconstructed geometries.\label{fig:Detecting-and-region}}
\end{figure*}

There still exist some noises in candidates. These noises include
\textit{irregular noises }and\textit{ regular noises}. Irregular noises
result from dramatic illumination variants or slight deviations between
frames. They may randomly appear in some consecutive frames. Generally,
it is not necessary to design an algorithm of pruning the irregular
noises since KF tracking can gradually eliminate this type of noise.
 In contrast, we term the background moving patterns as the regular
noises. Particularly, these noises are caused by the slight deviation
of satellite moving. Such deviation may be appeared/detected as the
edges or corners of some static objects in the frames. Even worse,
these detected corners or edges exhibit relative moving pattern with
respect to the moving background. The regular noises have to be pruned
by our algorithms. We visualize the regular noises in Fig. \ref{fig:Pseudo-motions}.

To this end, we propose a novel discrimination algorithm using the
geometrical and neighborhood information. This discrimination algorithm
includes two parts, \textit{i.e., }\textit{\emph{the}}\textit{ region
growing} to reconstruct candidate geometry and \textit{multi-morphological-cue
based discrimination }to distinguish noises from vehicles. The key
idea is that a vehicle target is a \textit{singular point} in 2D temporal
domain. By contrast, these regular noises share similar temporal distributions
as their neighborhood in frames. If the candidates can be connected
with their similar neighborhood pixels, we can differentiate vehicles
from regular noises in terms of shape: the vehicle targets approximate
a rectangle, while regular noises can be taken as arbitrary shapes. 

\noindent \textbf{(1) Region growing. }A region growing algorithm
is proposed to connect a candidate with its similar neighborhood pixels.
This procedure is namely to reconstruct the whole geometry of a candidate.
From Fig. \ref{fig:Detecting-and-region}, the detector only yields
a partial geometry of a candidate, because of the overlap of positions
of a candidate in two adjacent frames. The region growing utilizes
the detected partial geometry to restore the whole geometry of the
candidate. Neighborhood area is defined as a $11\times11$ pixel window
in the candidate. Gaussian distribution is employed to measure the
similarity between a neighbor pixel and the candidate. The CDF of
Gaussian distribution is 

\begin{equation}
c_{G}(x\lyxmathsym{\textSFxi}\mu,\sigma)=\frac{1}{2}\left[1+erf\left(\frac{x-\mu}{\text{\ensuremath{\sqrt{2}}}\sigma}\right)\right]\label{eq:6}
\end{equation}

\noindent where $erf(\cdot)$, $\text{\ensuremath{\mu}}$ and $\sigma$
denote the related error function, the mean and the standard deviation,
respectively. These parameter values of a Gaussian distribution can
be estimated using the values of those pixels of the candidate. A
range can be obtained given the predefined the lower bound probability
$p_{fa}^{-}$ and the upper bound probability $p_{fa}^{+}$, \emph{i.e}.
\begin{equation}
th_{G}^{-}=c_{G}^{-1}(p_{fa}{}^{-}\lyxmathsym{\textSFxi}\mu,\sigma)\label{eq:7}
\end{equation}

\begin{equation}
th_{G}^{+}=c_{G}^{-1}(p_{fa}{}^{+}\lyxmathsym{\textSFxi}\mu,\sigma)\label{eq:8}
\end{equation}
where $th_{G}^{-}$ and $th_{G}^{+}$ represent lower and upper bound
threshold, respectively. $p_{fa}^{-}$ and $p_{fa}^{+}$ are set as
$5\times10^{-3}$ and $1-5\times10^{-3}$ , symmetrically. If the
grey-level value of a pixel inner the searching window is in $\left[th_{G}^{-},th_{G}^{+}\right]$,
the pixel will be re-classified as candidate pixels. Those new candidate
pixels connected with the original candidate are reserved. Finally,
the result of region growing is shown in Fig. \ref{fig:Detecting-and-region}
(i) - (l). From Fig. \ref{fig:Detecting-and-region}, the detector
captures only partial geometries of the candidates in advance. Then,
the proposed region growing algorithm reconstruct the whole geometries
of the candidates. The region growing result also demonstrates the
feasibility of discrimination in terms of shape.

\begin{figure*}
\begin{centering}
\subfloat[]{\includegraphics{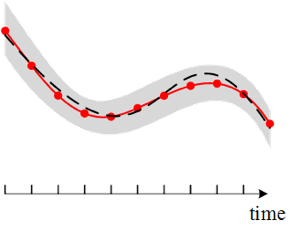}}\hspace{0.5cm}\subfloat[]{\includegraphics{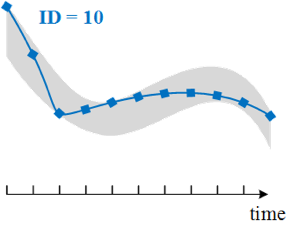}}
\par\end{centering}
\begin{centering}
\subfloat[]{\includegraphics{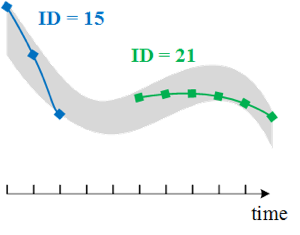}}\hspace{0.5cm}\subfloat[]{\includegraphics{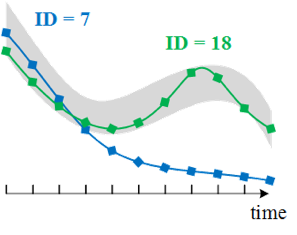}}
\par\end{centering}
\caption{Some cases of hypothesis-to-ground-truth associations. In panel (a),
the black dash line, the red solid line with filled circles, and the
grey filled polygon denote accurate trajectory of a vehicle target,
manual annotated ground truth of the trajectory, and the area where
an output belongs to the trajectory, respectively. (a) illustrates
that manual annotated ground truth cannot completely fit the accurate
trajectory. (b) shows a hypothesis whose trajectory fits the ground
truth in panel (a). (c) shows that two hypotheses cover the same ground
truth, wherein IDSW is counted. From panel (d), the hypothesis whose
ID equal to 7 outperforms the hypothesis whose ID equal to 18 during
first three frames. However, the former gradually loses the main pattern
of the ground truth, the latter follows the ground truth more closely.\label{fig:Some-cases-of}}
\end{figure*}

\noindent \textbf{(2) Multi-Morphological-Cue Based Discrimination.}
After region growing, we adopt a series of morphological properties
to differentiate vehicle targets and irregular noises. The employed
morphological cues include \textit{area, extent, major axis length}
and\textit{ eccentricity} as follows. 

\noindent \textbf{Area}. The number of the pixels of a candidate.

\noindent \textbf{Extent}. The ratio of pixels of a candidate to
the area of the bounding box of the candidate. 

\noindent \textbf{Major Axis Length}. If an ellipse has the same
normalized second central moments as the connected region of a candidate,
the major axis length of the ellipse is defined as the major axis
length of the candidate. 

\noindent \textbf{Eccentricity}. The eccentricity of a candidate
is equal to the eccentricity of the above ellipse.

The area and the major axis length cues represent the size of a candidate,
while the extent and the eccentricity cues measure the similarity
between a candidate and a rectangle. The spacing in the satellite
videos represents about 1 meter in real world. Thereby, these morphological
cues indicate real shape of a vehicle. They constitute a robust feature
of vehicles because vehicles are rigid bodies without any deformation
in satellite videos. 

\section{Metrics and Protocols in Performance Evaluation\label{sec:Metrics-and-Protocol}}

The widely-used evaluation metrics on object detection tasks are precision/recall
curve, and average precision (AP) \cite{everingham2010pascal_voc}.
These metrics are widely used in traditional visual object benchmarks,
\emph{i.e}. PASCAL VOC \cite{everingham2010pascal_voc} and MOT \cite{milan2016mot16}.
As a novel task of detecting tiny moving objects, these previous metrics
are inefficient in evaluating the performance of our task. The key
challenge again is caused by the fact that each vehicle has only several
pixels on each video frame. To this end, we systematically introduce
a complete set of evaluation protocol in measuring the algorithm performance
on our novel detection tasks in Sec. \ref{subsec:An-Evaluation-Protocol}.
Our evaluation protocol is built upon the existing evaluation metrics
in Sec. \ref{subsec:Evaluation-metrics}. 

\subsection{Evaluation Metrics\label{subsec:Evaluation-metrics}}

Generally, a single criterion cannot reckon the performance of detecting
and tracking objectively and comprehensively. To the best of our knowledge,
it is the first time to introduce a systematic series of evaluation
metrics, including precision, recall, Jaccard similarity, etc., whose
definitions detail as following.

\vspace{0.05in}

\noindent \textbf{Precision.} With respect to detection performance
evaluation, it is the most important to determine whether a hypothesis
is a true positive (TP) that is an accurate target correctly covered
by an output, or a false positive (FP) that is a non target falsely
covered by an output. Those missed accurate targets are called as
false negatives (FNs). The ratio of the accurate targets to the detected
targets is Precision, i.e.
\begin{equation}
Precision=\frac{TP}{TP+FP}\label{eq:18}
\end{equation}

\vspace{0.05in}

\noindent \textbf{Recall}. Recall measures the ability of a detector
to capture true target, which is equal to the ratio of TP to the number
of all existing true targets, namely

\begin{equation}
Recall=\frac{TP}{TP+FN}\label{eq:19}
\end{equation}

\vspace{0.05in}

\noindent \textbf{${\color{black}{\color{black}F}}_{{\color{black}{\color{black}1}}}$-score.
$F_{1}$}-score is a traditional criterion of binary classification
between interest targets and non targets, which is equal to the harmonic
mean of Precision and Recall, \emph{i.e}.

\begin{equation}
F_{1}=2\cdot\dfrac{Precision\cdot Recall}{Precision+Recall}\label{eq:23}
\end{equation}

\vspace{0.05in}

\noindent \textbf{Jaccard Similarity}. Jaccard similarity is a criterion
of evaluating tracking performance, which integrates TP, FP and FN
as follows, \cite{liang2014novel} 

\begin{equation}
J=\frac{TP}{TP+FP+FN}\label{eq:20}
\end{equation}

\vspace{0.05in}

\begin{table*}
\begin{centering}
\begin{tabular}{c|c|c|c|c|c|c|c|c}
\hline 
\multirow{2}{*}{{\scriptsize{}District }} & \multirow{2}{*}{{\scriptsize{}Frame Rate (fps) }} & \multirow{2}{*}{{\scriptsize{}Resolution (m)}} & \multirow{2}{*}{{\scriptsize{}Duration (s) }} & \multirow{2}{*}{{\scriptsize{}Height\texttimes Width (pixel)}} & \multicolumn{4}{c}{{\scriptsize{}Latitude and Longitude of Frame Corner}}\tabularnewline
\cline{6-9} 
 &  &  &  &  & {\scriptsize{}Top Left } & {\scriptsize{}Top Right } & {\scriptsize{}Bottom Left } & {\scriptsize{}Bottom Right}\tabularnewline
\hline 
\hline 
{\scriptsize{}Valencia, Spain} & {\scriptsize{}20} & {\scriptsize{}1.0} & {\scriptsize{}29} & {\scriptsize{}3072\texttimes 4096 } & {\scriptsize{}$\begin{array}{c}
39.4989N\\
0.3719W
\end{array}$ } & {\scriptsize{}$\begin{array}{c}
39.4928N\\
0.3278W
\end{array}$ } & {\scriptsize{}$\begin{array}{c}
29.4731N\\
0.3775W
\end{array}$ } & {\scriptsize{}$\begin{array}{c}
39.4669N\\
0.3333W
\end{array}$ }\tabularnewline
\hline 
\end{tabular}
\par\end{centering}
\caption{Information of the experimental satellite video.\label{tab:Information-of-experimental} }
\end{table*}

\noindent \textbf{MOTA}. Multiple object tracking accuracy (MOTA)
is a tracking performance metric to quantify multiple object tracking
performance. The definition of MOTA is \cite{milan2016mot16,kasturi2009framework}

\begin{equation}
MOTA=1-\frac{\sum_{i}\left(FN_{i}+FP_{i}+IDSW_{i}\right)}{\sum_{i}GT_{i}}\label{eq:21}
\end{equation}

\noindent where $FN_{i}$, $FP_{i}$, $IDSW_{i}$ and $GT_{i}$ represent
the number of $FN$, $FP$, $IDSW$ and ground truth, respectively,
in frame i. IDSW means identity switch of the trajectories associated
to a ground truth, and please refer to \cite{milan2016mot16} for
more details. Obviously, MOTA score ranges from -\ensuremath{\infty}
to 1, and the bigger it is, the better the goodness of detecting and
tracking is.

\vspace{0.05in}

\begin{figure*}
\begin{centering}
\subfloat[]{\includegraphics{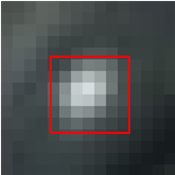}}\hspace{0.5cm}\subfloat[]{\includegraphics{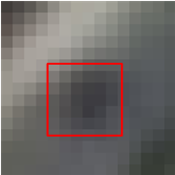}}\hspace{0.5cm}\subfloat[]{\includegraphics{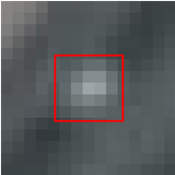}}\hspace{0.5cm}\subfloat[]{\includegraphics{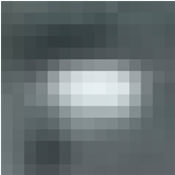}}
\par\end{centering}
\begin{centering}
\subfloat[]{\includegraphics{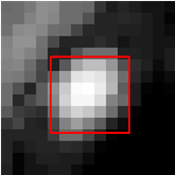}}\hspace{0.5cm}\subfloat[]{\includegraphics{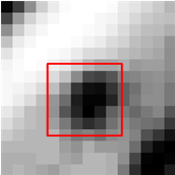}}\hspace{0.5cm}\subfloat[]{\includegraphics{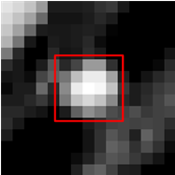}}\hspace{0.5cm}\subfloat[]{\includegraphics{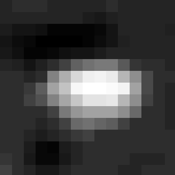}}
\par\end{centering}
\caption{Some vehicle samples and a noise. These images are scaled for viewing.
(a), (b), (c) and (d) are RGB images where red rectangles represent
real vehicles, while a noise signal exists in image (d). Greyscale
images (e), (f), (g) and (h) are corresponding enhanced images of
(a), (b), (c) and (d) to improve contrast for viewing. \label{fig:Some-vehicle-samples}}
\end{figure*}

\noindent \textbf{MOTP}. Multiple object tracking precision (MOTP)
is adopted to measure the positioning precision of the detecting algorithms,
which can be written as \cite{milan2016mot16,kasturi2009framework,stiefelhagen2006clear}

\begin{equation}
MOTP=\frac{\sum_{i}IoU_{i}}{\text{\ensuremath{\sum}}_{i}M_{i}}\label{eq:22}
\end{equation}

\noindent where $IoU_{i}$ (intersection over union \cite{everingham2010pascal_voc})
denote the sum of overlap ratio of hypotheses to ground truths and
$M_{i}$ is the number of matches of ground truths and hypotheses
in frame \textit{i}. From the above definition, MOTP score ranges
from 0 to 1, and the bigger that is, the more precise the derived
location is.

\begin{figure}
\centering{}\includegraphics[scale=0.5]{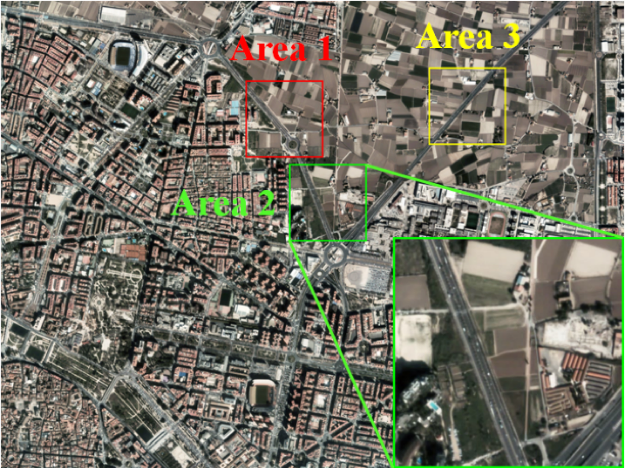}\caption{Annotated areas of the satellite video.\label{fig:Annotated-areas-of}}
\end{figure}

\subsection{An Evaluation Protocol\label{subsec:An-Evaluation-Protocol}}

The IoU of bboxes between a hypothesis and a ground truth is adopted
as the similarity between a hypothesis and a truth. Similar to aforementioned
\textcolor{black}{hypothesis-to-track association}, the matching of
multiple hypotheses and ground truths also resorts to \textcolor{black}{Hungary
algorithm \cite{Miller1997Hungary} }in spatiotemporal domain not
only in each frame. 

Some cases of associations are listed in Fig. \ref{fig:Some-cases-of}
(b)-(d), and actual situation is more complicated than that. The protocol
of performance evaluation of the proposed detection and tracking algorithm
details as following: 
\begin{enumerate}
\item The IoUs of each hypothesis and each ground truth can be obtained.Then,
the reci\textcolor{black}{procal of a IoU is the }\textit{\textcolor{black}{distance}}\textcolor{black}{{}
between a hypothesis and a ground truth. Note that all IoUs are added
with a very small value to avoid zero denominators. The matching distance
threshold is set as 50, empirically, which is equal to a very small
IoU value, 0.02. It means that if a hypothesis and a ground truth
overlap, the hypothesis is regarded to cover the ground truth. In
contrast, the IoU ratio is set as 0.5 for the detection of general
objects, such as pedestrian, aeroplane, bicycle, }\textcolor{black}{\emph{etc}}\textcolor{black}{.,
which cannot adopt to the tiny vehicles in satellite videos. Obviously,
the smaller the distance, the smaller is the cost of the association
between the hypothesis and the ground truth. All distances constitute
a cost matrix, given }\textit{\textcolor{black}{M}}\textcolor{black}{{}
hypotheses and }\textit{\textcolor{black}{N }}\textcolor{black}{ground
truths, i.e.
\begin{equation}
CM_{t}=\left[\begin{array}{cccc}
\frac{1}{IoU_{1,1}} & \frac{1}{IoU_{1,2}} & \ldots & \frac{1}{IoU_{1,N}}\\
\frac{1}{IoU_{2,1}} & \frac{1}{IoU_{2,2}} & \ldots & \frac{1}{IoU_{2,N}}\\
\vdots & \vdots & \ddots & \vdots\\
\frac{1}{IoU_{M,1}} & \frac{1}{IoU_{M,2}} & \ldots & \frac{1}{IoU_{M,N}}
\end{array}\right]\label{eq:24}
\end{equation}
where }\textit{\textcolor{black}{t}}\textcolor{black}{{} indicates the
frame number.}
\item \textcolor{black}{Repeat the first step for }\textit{\textcolor{black}{K}}\textcolor{black}{{}
consecutive frames. Then, cost matrices of these frames can constritue
a cost tensor, namely
\begin{equation}
CT=\left[CM_{1},CM_{2},\cdots,CM_{K}\right]\label{eq:25}
\end{equation}
}
\item \textcolor{black}{The optimal associations of hypotheses and ground
truths can be obtained using Hungary algorithm. A time win}dow is
employed to reduce computational load and memory consumption. Explicitly,
the association is implemented among ten consecutive frames rather
than all frames, namely the \textit{K} in Eq (\ref{eq:25}) is set
as 10.
\item Repeat the step 1-3.
\item Finally, the metrics are calculated based on the associations. 
\end{enumerate}

\section{Experiments and Discussion\label{sec:Experiments-and-Discussion}}

\subsection{Experimental Setups}

\begin{figure}
\begin{centering}
\subfloat[]{\includegraphics[scale=0.7]{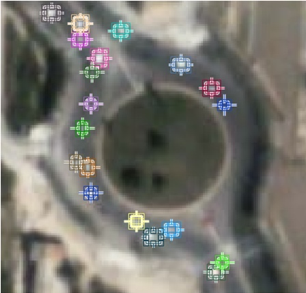}}\hspace{0.5cm}\subfloat[]{\includegraphics[scale=0.7]{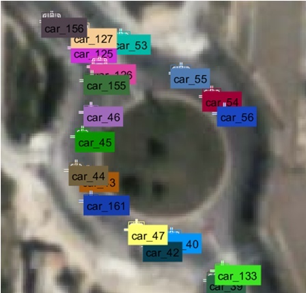}}
\par\end{centering}
\caption{One shot of ground truth annotation. (a) shows the locations of vehicles,
while (b) represents their corresponding IDs.\label{fig:One-shot-of}}
\end{figure}

\begin{figure*}
\begin{centering}
\subfloat[Frame 50]{\includegraphics{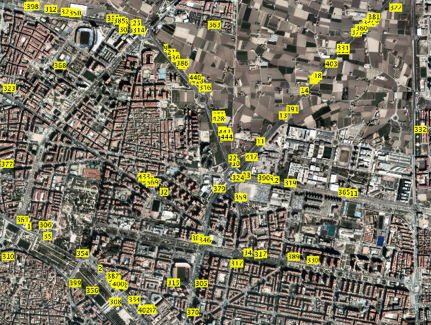}}\hspace{0.5cm}\subfloat[Frame 100]{\includegraphics{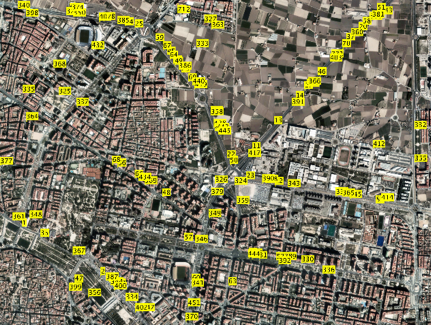}}
\par\end{centering}
\begin{centering}
\subfloat[Frame 150]{\includegraphics{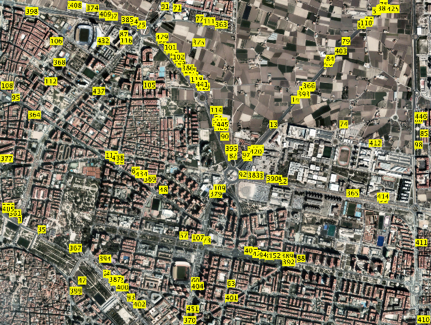}}\hspace{0.5cm}\subfloat[Frame 200]{\includegraphics{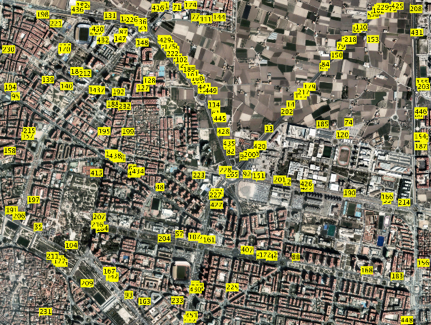}}
\par\end{centering}
\caption{Tiny vehicle detecting results of four frames. Please refer to the
enlargements of scenes in Fig. \ref{fig:The-enlargement-of-1} \label{fig:Tiny-vehicle-detecting} }
\end{figure*}

\noindent \textbf{Dataset}. The experimental satellite video is provided
by CGSTL. The videos are captured on March 7, 2017. A video satellite
recorded a region in Valencia, Spain. The information of the satellite
video details as Table \ref{tab:Information-of-experimental}. The
study video is free provided by CGSTL for scientific research.\footnote{ Any one can purchases satellite videos from their official website\url{http://mall.charmingglobe.com/videoIndex.html}}
The annotated dataset can be downloaded from the first authour's website.

\noindent \textbf{Competitors}. We compare several other algorithms
in detecting the tiny moving vehicles, including GMM and ViBe. In
order to fairly compare the proposed algorithm with baseline algorithms,
GMM and ViBe are also linked with the same KF tracking framework. 

\begin{figure}
\begin{centering}
\subfloat[ID = 19 ]{\includegraphics{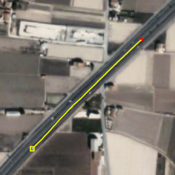}}\hspace{0.5cm}\subfloat[ID = 359 ]{\includegraphics{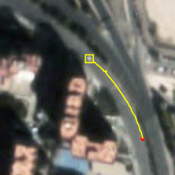}}\hspace{0.5cm}\subfloat[ID = 92]{\includegraphics{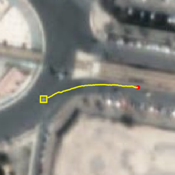}}\hspace{0.5cm}\subfloat[ID = 161 ]{\includegraphics{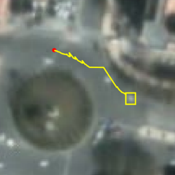}}
\par\end{centering}
\caption{The tracks of four vehicles. The yellow lines indicate their moving
tracks detected by our algorithms. The yellow rectangles mark their
initial positions, while the filled red points denote their terminal
locations. In contrast, the other two methods \textendash{} GMM and
ViBe can not be used to detecting/tracking the vehicles. \label{fig:The-tracks-of}}
\end{figure}

\begin{figure*}
\begin{centering}
\subfloat[]{\includegraphics[scale=0.7]{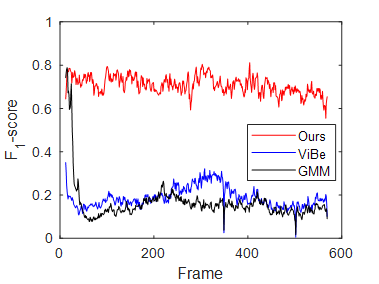}}\hspace{0.5cm}\subfloat[]{\includegraphics[scale=0.7]{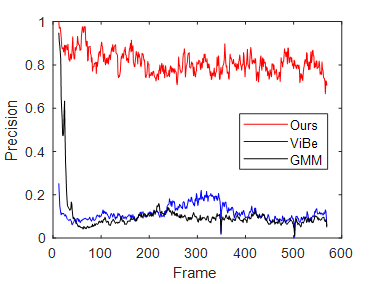}}\hspace{0.5cm}\subfloat[]{\includegraphics[scale=0.7]{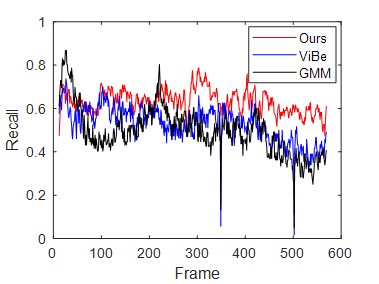}}
\par\end{centering}
\caption{Detecting scores of the proposed algorithm and baselines frame by
frame.\label{fig:Detection-and-tracking}}
\end{figure*}

\begin{figure*}
\begin{centering}
\subfloat[]{\includegraphics[scale=0.035]{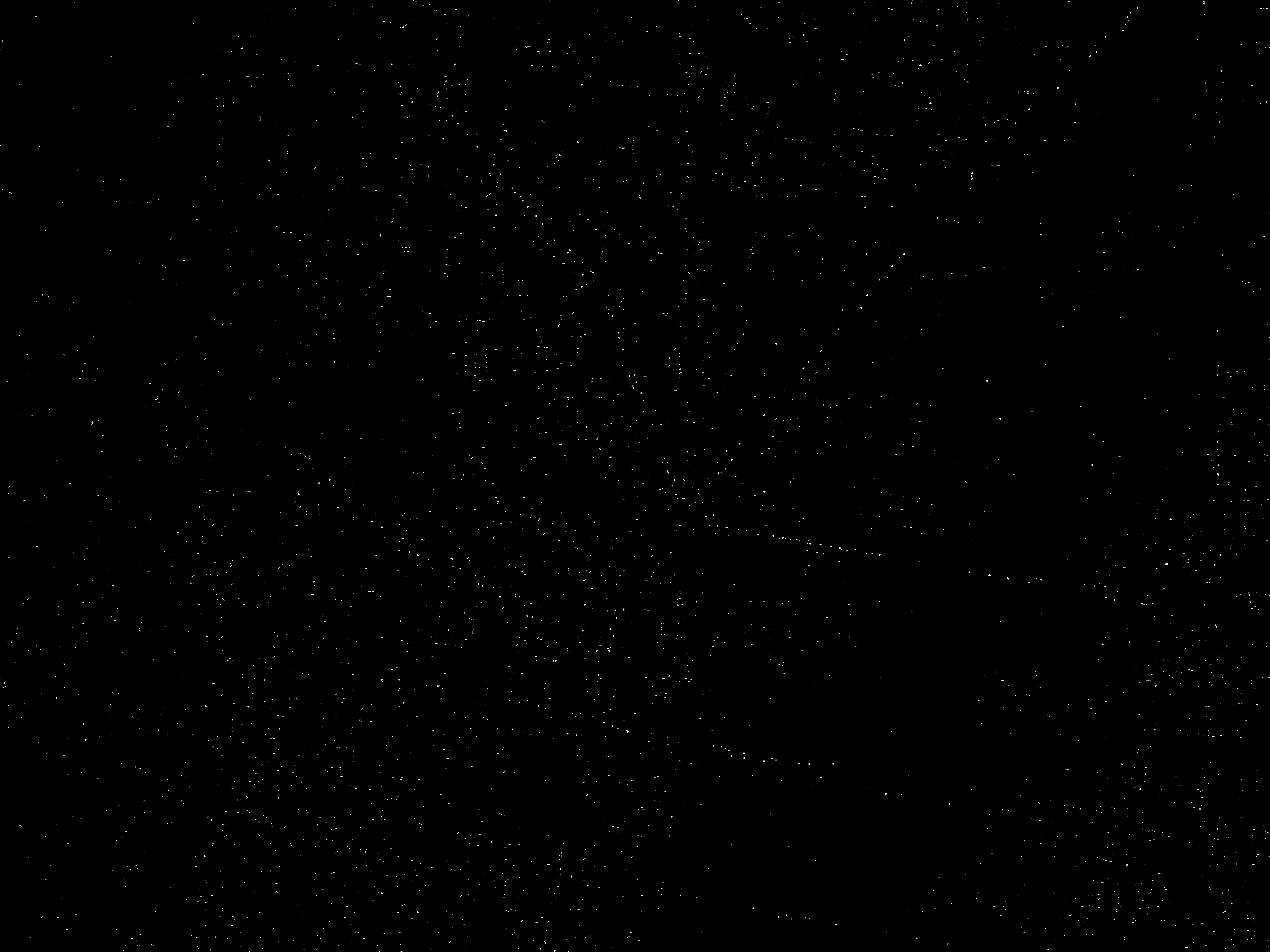}}\hspace{0.5cm}\subfloat[]{\includegraphics[scale=0.035]{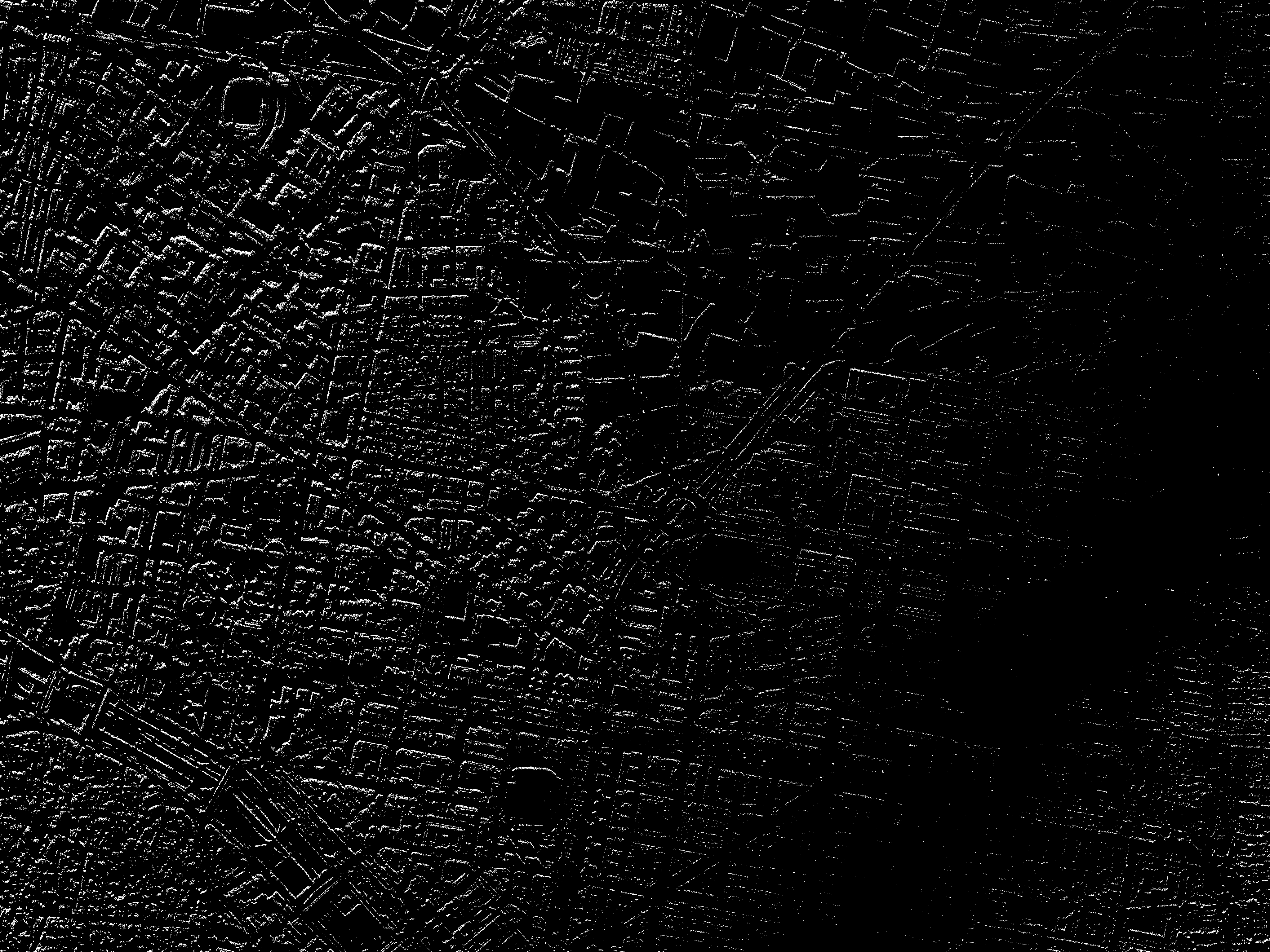}}\hspace{0.5cm}\subfloat[]{\includegraphics[scale=0.035]{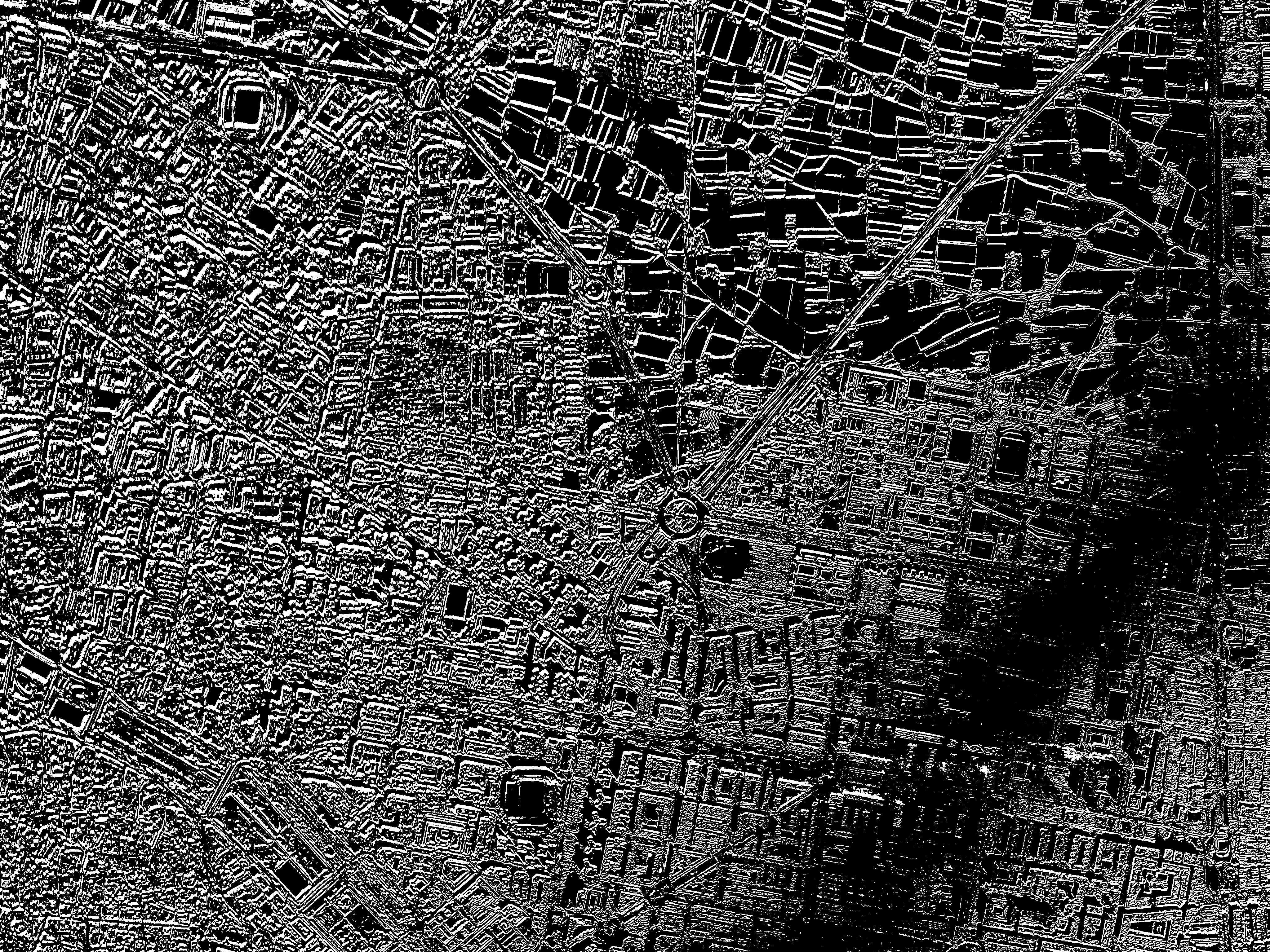}}
\par\end{centering}
\caption{Foreground segmentation results of Frame 100. (a) the proposed algorithm,
(b) ViBe, (c) GMM.\label{fig:Foreground-segmentation-results}}
\end{figure*}

\noindent \textbf{Ground-truth}. We annotated the experimental satellite
video to quantitatively evaluate the proposed algorithm. The annotation
of satellite videos is an arduous work, because the vehicles are hard
to be distinguished from the background in naked eyes for a lack of
distinctive prominent features, which is illustrated in Fig. \ref{fig:Some-vehicle-samples}.
In Fig. \ref{fig:Some-vehicle-samples} (a), (b) and (c), there are
no significant differences between real vehicle targets and the background,
especially dark vehicles, like Fig. \ref{fig:Some-vehicle-samples}
(b). Fig. \ref{fig:Some-vehicle-samples} (d) shows that a noise signal
may be a stationary vehicle, and resembles the true positive, which
results in ambiguities in the interpreting work. In order to tackle
this problem, we inspect the previous and future frames to determine
whether some connected pixels form a moving vehicle. In the other
words, we go through a short-term consecutive frames to seek out a
moving region. Besides the difficult to detect vehicles, we can hardly
annotate all vehicles of the scope spanning about 3\texttimes 4 square
kilometers frame by frame. We also seek the trade-off between workload
and annotation accuracy here: first, three areas with 500\texttimes 500
pixels of the video are randomly selected to be annotated, as illustrated
in Fig. \ref{fig:Annotated-areas-of}; second, we manually annotate
the vehicles every 10 frames, while the ground-truth of the other
frames are obtained by linear interpolation. Fig. \ref{fig:One-shot-of}
shows a representative scene of the annotation. The vehicle numbers
of ground truths of area 1, area 2 and area 3 are 49, 41 and 29, respectively.
Particularly, we utilized the Ground Truth Labeler app in MATLAB 2018a
to help annotate the satellite video.

\subsection{Experimental Results and Discussion}

\begin{figure*}
\subfloat[]{\includegraphics[scale=0.6]{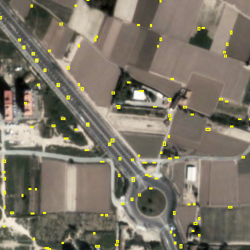}}\hspace{0.5cm}\subfloat[]{\includegraphics[scale=0.6]{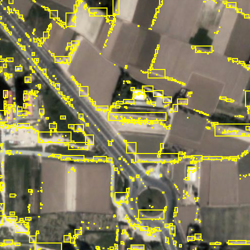}}\hspace{0.5cm}\subfloat[]{\includegraphics[scale=0.6]{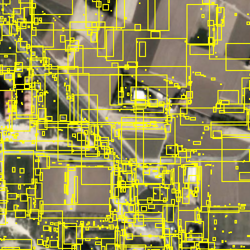}}

\subfloat[]{\includegraphics[scale=0.6]{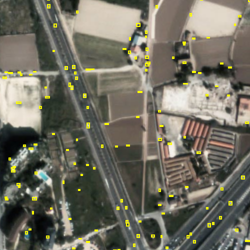}}\hspace{0.5cm}\subfloat[]{\includegraphics[scale=0.6]{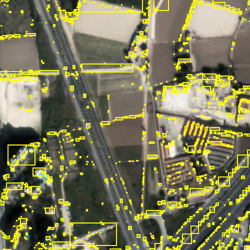}}\hspace{0.5cm}\subfloat[]{\includegraphics[scale=0.6]{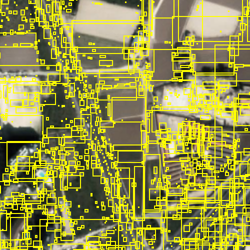}}

\subfloat[]{\includegraphics[scale=0.6]{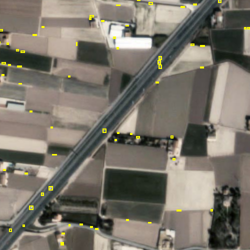}}\hspace{0.5cm}\subfloat[]{\includegraphics[scale=0.6]{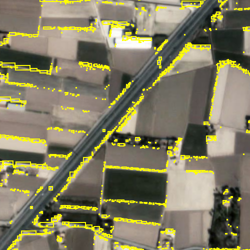}}\hspace{0.5cm}\subfloat[]{\includegraphics[scale=0.6]{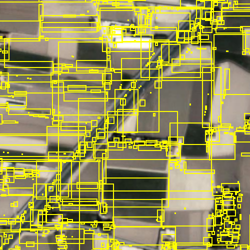}}\caption{Vehicle candidates yielded by our method, ViBe and GMM in frame 100.
The first, second, and third column represent the results produced
by our method, ViBe and GMM, respectively. The first, second and third
row are corresponding to the vehicle candidates in Area 1, Area 2,
Area 3, respectively.\label{fig:Vehicle-candidates-yielded}}
\end{figure*}

\noindent \textbf{Quantitative results. }The quantitative results
of comparison experiments are detailed in Table \ref{tab:Evaluation-scores-of},
including separate areas and average results. Table \ref{tab:Evaluation-scores-of}
shows that the proposed framework obviously outperforms ViBe and GMM
in any criterion. Specifically, ViBe and GMM detect around 50\% of
vehicles, but yield about 90\% of FPs. The high false positive rate
extremely degrades the performance of ViBe and GMM. That results from
that ViBe and GMM totally neglects the background moving and also
fails to separate moving vehicles from noises. On the other hand,
they are both sensitive the varying of the pixels, so, they report
a trivial Recall score: 50\%. On the contrary, our method has the
unique capability of perceiving pixel \textit{moving} not only\textit{
varying}. Leveraging the very high Precision, our method reports good
scores of $F_{1}$-score, Jaccard Similarity and MOTA. These criteria
are closely related to Precision. On the other hand, the Recall score
of our method is relatively low, although it is about 10\% higher
than ViBe and GMM. Some related criteria, Jaccard Similarity and MOTA,
is affected to some extent. The location precision metric, MOTP, shows
that average overlap between ground truths and hypotheses is 0.52.
The very small size of vehicle targets leads to difficult in precisely
locating them. But we think the location precision and pixel-level
deviation meet the demand of application, especially in urban traffic
surveillance. 

\noindent \textbf{Qualitative results. }We give some visualization
results of detecting multiple tiny moving vehicles. Figure \ref{fig:Tiny-vehicle-detecting}
shows four frames, frame 50, 100, 150, 200. Intuitively, many vehicles
in the main arteries are detected by the proposed algorithms, and
there are a few FPs because most of the annotated labels exist in
the main arteries. On the other hand, the number of detected vehicles
is growing gradually frame by frame, which illustrates that the tracking
also facilitates the detecting. 

\begin{figure*}
\begin{centering}
\subfloat[]{\includegraphics{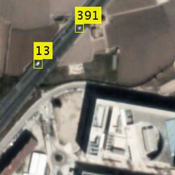}}\hspace{0.5cm}\subfloat[]{\includegraphics{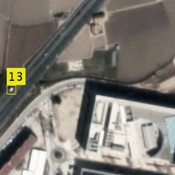}}\hspace{0.5cm}\subfloat[]{\includegraphics{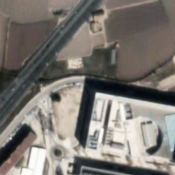}}\hspace{0.5cm}\subfloat[]{\includegraphics{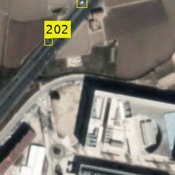}}
\par\end{centering}
\begin{centering}
\subfloat[]{\includegraphics{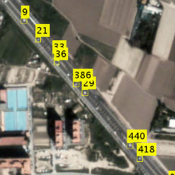}}\hspace{0.5cm}\subfloat[]{\includegraphics{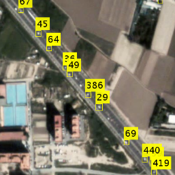}}\hspace{0.5cm}\subfloat[]{\includegraphics{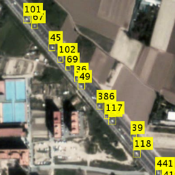}}\hspace{0.5cm}\subfloat[]{\includegraphics{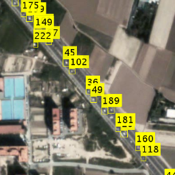}}
\par\end{centering}
\begin{centering}
\subfloat[]{\includegraphics{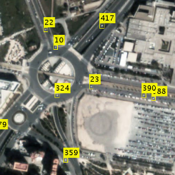}}\hspace{0.5cm}\subfloat[]{\includegraphics{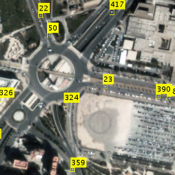}}\hspace{0.5cm}\subfloat[]{\includegraphics{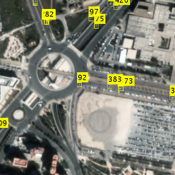}}\hspace{0.5cm}\subfloat[]{\includegraphics{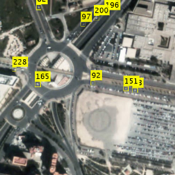}}
\par\end{centering}
\centering{}\subfloat[]{\includegraphics{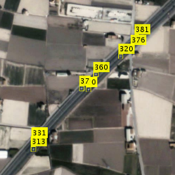}}\hspace{0.5cm}\subfloat[]{\includegraphics{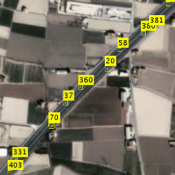}}\hspace{0.5cm}\subfloat[]{\includegraphics{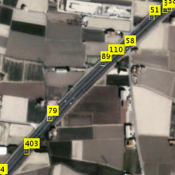}}\hspace{0.5cm}\subfloat[]{\includegraphics{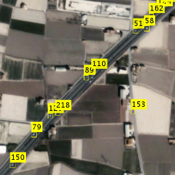}}\caption{The enlargements of four scenes of Fig. \ref{fig:Tiny-vehicle-detecting}.
\label{fig:The-enlargement-of-1}}
\end{figure*}

In order to clearly observe the detecting and tracking details, we
provide four enlargements as shown in Fig. \ref{fig:The-enlargement-of-1}.
From the Fig. \ref{fig:The-enlargement-of-1}, we can obtain the dynamics
of not only the moving of vehicles but also the detecting and tracking
procedure. Once a vehicle has been repeatedly detected of two consecutive
frames, a unique ID is assigned to it and simultaneously a KF is allotted
to it. The position and velocity provided by the detector also are
used to initialize the KF. Subsequently, according to the designed
framework, the detector provides the current state frame by frame,
while the KF continuously embodies the latest state of the tracking
vehicle and further updates its systematic model. Therefore, the proposed
processing workflow can detect the tiny moving vehicles accurately
and precisely. 

Fig. \ref{fig:The-tracks-of} shows four trajectories tracked by the
proposed algorithms. These trajectories include linear tracks and
curved tracks, which demonstrates the above theory that a series of
linear procedures can approximate a non-linear procedure as accurate
as possible. These four trajectories also cover different traffic
scenarios, including a straight artery in Fig. \ref{fig:The-tracks-of}(a),
a right turn in Fig. \ref{fig:The-tracks-of}(b) and two roundabouts
in Fig. \ref{fig:The-tracks-of}(c)-(d). It proves that the proposed
algorithms can not only address simple traffic conditions but also
adapt to complex traffic scenarios. 

\begin{table}
\begin{centering}
\begin{tabular}{c|c|c|c|c|c|c|c}
\hline 
Area & Method & R. (\%)  & P. (\%) & $F_{1}$ & JS & MA  & MP \tabularnewline
\hline 
\hline 
\multirow{3}{*}{1} & Ours & 64.15 & 81.71 & 0.72 & 0.56 & 0.46 & 0.50\tabularnewline
\cline{2-8} 
 & ViBe & 51.72 & 15.10 & 0.23 & 0.13 & -2.45 & 0.39\tabularnewline
\cline{2-8} 
 & GMM & 43.82 & 12.29 & 0.19 & 0.11 & -2.75 & 0.37\tabularnewline
\hline 
\multirow{3}{*}{2} & Ours & 62.80 & 82.23 & 0.71 & 0.55 & 0.47 & 0.52\tabularnewline
\cline{2-8} 
 & ViBe & 61.70 & 9.14 & 0.16 & 0.09 & -5.56 & 0.45\tabularnewline
\cline{2-8} 
 & GMM & 61.83 & 7.5 & 0.13 & 0.07 & -7.08 & 0.39\tabularnewline
\hline 
\multirow{3}{*}{3} & Ours & 60.42 & 77.26 & 0.68 & 0.51 & 0.41 & 0.56\tabularnewline
\cline{2-8} 
 & ViBe & 41.53 & 6.76 & 0.12 & 0.06 & -5.35 & 0.47\tabularnewline
\cline{2-8} 
 & GMM & 46.10 & 6.34 & 0.11 & 0.06 & -6.41 & 0.42\tabularnewline
\hline 
\multirow{3}{*}{Avg.} & Ours & 63.06 & 81.04 & 0.71 & 0.55 & 0.46 & 0.52\tabularnewline
\cline{2-8} 
 & ViBe & 52.86 & 10.74 & 0.18 & 0.10 & -3.92 & 0.43\tabularnewline
\cline{2-8} 
 & GMM & 49.66 & 8.79 & 0.15 & 0.08 & -4.72 & 0.39\tabularnewline
\hline 
\end{tabular}
\par\end{centering}
\caption{Evaluation scores of the proposed algorithm and baseline algorithms.
R., P., $F_{1}$, JS, MA and MP are short for Recall, Precision, $F_{1}$-score,
Jaccard Similarity, MOTA and MOTP.\label{tab:Evaluation-scores-of}}
\end{table}

Figure \ref{fig:Foreground-segmentation-results} shows the foreground
segmentation results generated by our algorithms and baseline algorithms.
From Fig. \ref{fig:Foreground-segmentation-results} (a), our detector
yields hypotheses composed of true vehicles and a few noise, while
the noise extremely outnumber true vehicles in \ref{fig:Foreground-segmentation-results}
(b) and (c). The candidate vehicles generated by the proposed algorithm
and competitors area shown in Fig. \ref{fig:Vehicle-candidates-yielded}.
From Fig. \ref{fig:Vehicle-candidates-yielded}, our detector mainly
perceives the moving vehicle pixels in the roads and yields limited
false positives, while ViBe and GMM aimlessly detect varying of the
pixels. It illustrates that ViBe and GMM are unable to separate the
motions of vehicles from the slow and slight motion of background.
ViBe and GMM try to estimate the pattern of each pixel using a non-parametric
model or a Gaussian distribution, respectively. So, this strategy
is very sensitive to the varying of the pixel and works well in common
video processing, especially a stationary camera. However, they cannot
address the satellite video processing because of neglecting local
or neighborhood information. Our detection algorithms focus on a local
area not a single pixel, which can adapt to the moving background
of satellite videos.

\noindent \textbf{Frame-by-frame Quantitative results. }For providing
a detailed figure of detecting performance of the proposed algorithm
and baselines, some scores frame by frame are presented in Fig. \ref{fig:Detection-and-tracking}.
Fig. \ref{fig:Detection-and-tracking} shows the most widely-used
detecting metrics: F\textsubscript{\textcolor{black}{1}}-score, Precision
and Recall. It further illustrates that the proposed algorithm outperforms
baselines mainly by leveraging high Precision. From Fig. \ref{fig:Detection-and-tracking},
the F\textsubscript{\textcolor{black}{1}}-score of ViBe and GMM rapidly
decays at the beginning and then totally traps into the moving background.
So, for ViBe and GMM, the background moving totally blurs the vehicle
moving, leading to their failure. Likewise, this experiment demonstrates
the outstanding performance of the proposed algorithm to separate
the background moving and the vehicle moving.

\section{Conclusion\label{sec:Conclusion}}

Satellite videos have the unique capability of observing a city-scale
region. This paper addresses the tiny vehicle detecting algorithm
in satellite videos, and we design the practical detecting and tracking
framework of tiny moving vehicles in satellite videos. It is the first
time to adopt a probabilistic distribution to represent the patterns
of noises in spatiotemporal domain, which facilitates us to differentiate
candidates from noise. We further propose the multi-morphological-cue
based discrimination algorithm to distinguish true vehicle targets
from a few existing noise. Another important issue is to introduce
a series of evaluation metrics and to propose a complete evaluation
protocol. The proposed algorithms are tested in three manual annotated
areas of a satellite video, which are also compared with baseline
algorithms. These experiments demonstrate the good performance of
our algorithms.

\section{Acknowledgment }

We are grateful to CGSTL for providing the satellite video data used
in this study.

\end{document}